\documentclass[conference]{IEEEtran}
\usepackage{times}

\usepackage[numbers]{natbib}
\usepackage{multicol}
\usepackage[bookmarks=true]{hyperref}
\usepackage{graphicx}
\usepackage{multirow}
\usepackage{caption}
\usepackage{todonotes}
\usepackage{subfiles} 
\usepackage[export]{adjustbox}
\usepackage{amsmath}
\usepackage{amssymb}
\usepackage{tikz}
\usepackage{subcaption}
\usepackage{amsfonts}
\usepackage[font=small,labelfont=bf]{caption}

\begin{document}

\title{Action Conditioned Tactile Prediction: case study on slip prediction}

\author{Author Names Omitted for Anonymous Review. Paper-ID 106}

\author{\authorblockN{Willow Mandil}
\authorblockA{School of Computer Science\\
University of Lincoln\\
Brayford, Lincoln, UK\\
18710370@students.lincoln.ac.uk}
\and
\authorblockN{Kiyanoush Nazari}
\authorblockA{School of Computer Science\\
University of Lincoln\\
Brayford, Lincoln, UK\\
19749684@students.lincoln.ac.uk}
\and
\authorblockN{Amir Ghalamzan E.}
\authorblockA{Lincoln Institute for Agri-food Technology\\ 
University of Lincoln\\
Brayford, Lincoln, UK}
aghalamzanesfahani@lincoln.ac.uk
}


%

\maketitle

\begin{abstract}

    Tactile predictive models can be useful across several robotic manipulation tasks, e.g. robotic pushing, robotic grasping, slip avoidance, and in-hand manipulation. However, available tactile prediction models are mostly studied for image-based tactile sensors and there is no comparison study indicating the best performing models. In this paper, we presented two novel data-driven action-conditioned models for predicting tactile signals during real-world physical robot interaction tasks (1) action condition tactile prediction and (2) action conditioned tactile-video prediction models. We use a magnetic-based tactile sensor that is challenging to analyse and test state-of-the-art predictive models and the only existing bespoke tactile prediction model. We compare the performance of these models with those of our proposed models. We perform the comparison study using our novel tactile enabled dataset containing 51,000 tactile frames of a real-world robotic manipulation task with 11 flat-surfaced household objects. Our experimental results demonstrate the superiority of our proposed tactile prediction models in terms of qualitative, quantitative and slip prediction scores.

\end{abstract}

\IEEEpeerreviewmaketitle

\section{Introduction}
    Humans use tactile sensation to understand physical properties, helping to develop a cause-effect understanding of the scene and use it to plan interactive actions. Tactile sensation is essential for building physical interaction perception~\cite{johansson2009coding, nazari2021tactile}. Within the robotics community, tactile sensation has been used for slip detection~\cite{chen2018tactile}. These reactive systems use high-frequency tactile sensors to adjust grip force, preventing object slippage~\cite{yi2018biomimetic}. However, reactive tactile systems are a limited use of this sensor modality. Human tactile cognition~\cite{nicholas2010active} helps with a series of interactive tasks, e.g. robust grasping and manipulating an object, in-hand manipulation, tactile exploration, pushing, compliant tasks like wiping a board or writing. Humans use predictive cognition~\cite{thoroughman2000learning, tseng2007sensory} to perform such complex manipulation tasks. We present tactile predictive models (\textbf{TPM}) that can be helpful across different interactive tasks via predictive control.

    Related works are very application focused and typically use vision based tactile sensors. For instance,~\citet{zhang2020towards} used a long-short term memory (\textbf{LSTM}) based recurrent neural network (\textbf{RNN}) within a larger slip prediction model. \citet{tian2019manipulation} used a video prediction based TPM that enabled a very simple manipulation task (using model predictive control) of single objects through tactile sensation. The existing works perform no exploration of TPMs performance. To address this, we tested and compared data-driven models performance in predicting tactile signals. Moreover, we propose two novel action-conditioned TPMs outperforming the existing approaches. We demonstrate the superiority of our proposed methods across several real robot household objects manipulation tasks using tactile sensors with sparse point-wise force measurements. The primary contributions of this paper are:

        \begin{figure}[t]
            \centering
            \adjustbox{fbox= 0pt 4pt, frame}{\includegraphics[trim=200pt 100pt 400pt 0pt, clip, width=0.945\columnwidth]{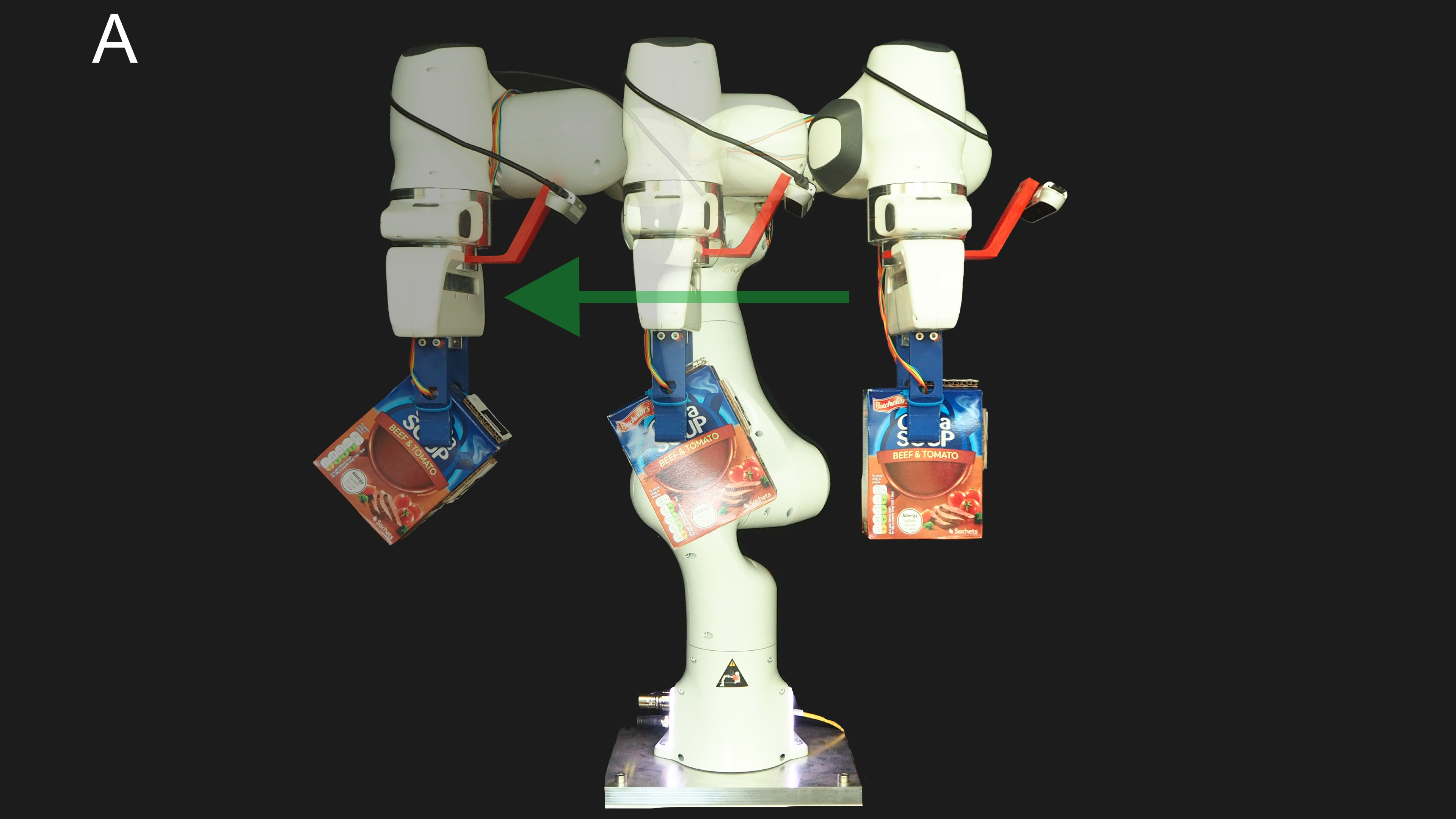}}

            \vspace{0.2cm}

            \adjustbox{fbox= 0pt 4pt, frame}{\includegraphics[trim=600pt 1000pt 600pt 600pt, clip, width=0.2\columnwidth]{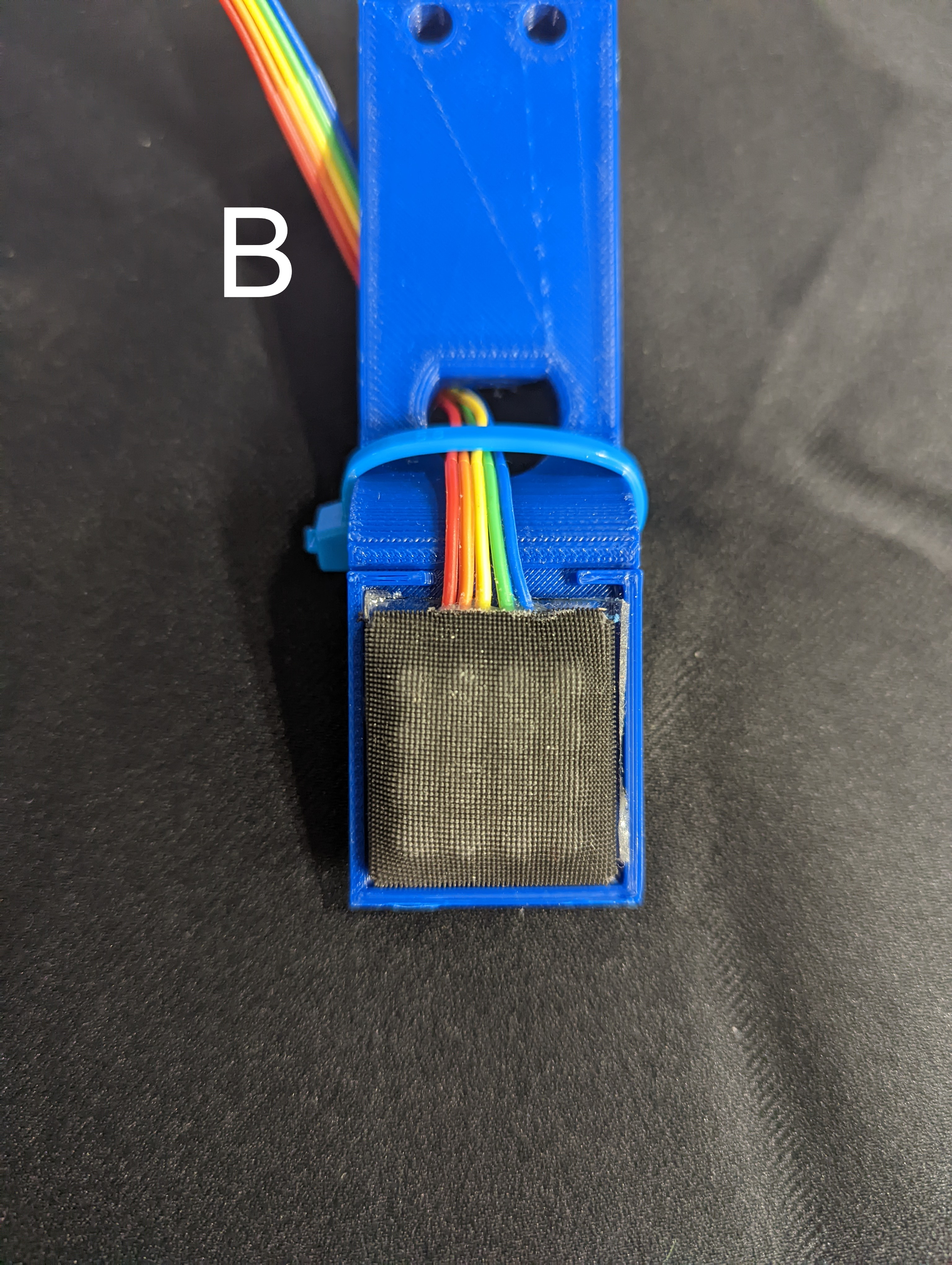}}
            \adjustbox{fbox= 0pt 4pt, frame}{\includegraphics[trim=0pt 0pt 200pt 0pt, clip, width=0.34\textwidth]{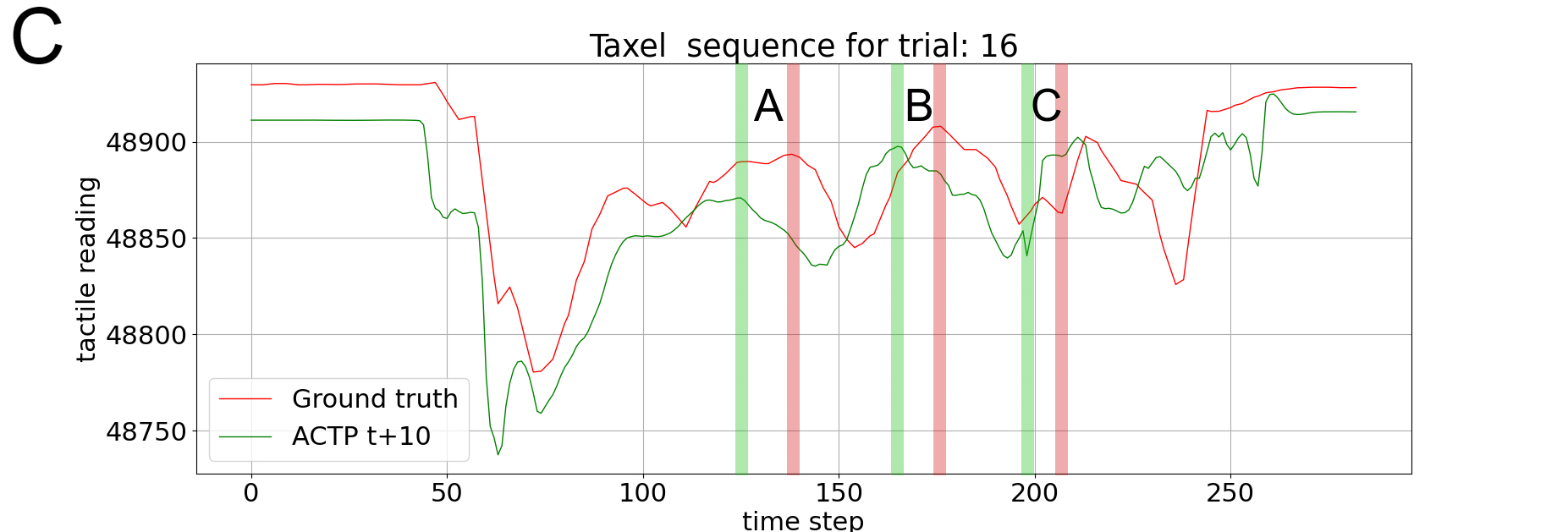}}
            \caption{(A) Teleoperated kinesthetic data collection with tactile finger tipped robot (B) Xela uSkin tactile sensor (C) Single taxel value during pick and move trial and ACTP tactile signal prediction. Letters and vertical bars indicating correct peak and trough predictions ahead of time.}
            \label{fig::intro_image}
        \end{figure}

        \begin{figure*}[t]
            \adjustbox{fbox= 0pt 4pt, frame}{
                \includegraphics[trim=0pt 0pt 0pt 0pt, clip, height=0.23\textwidth]{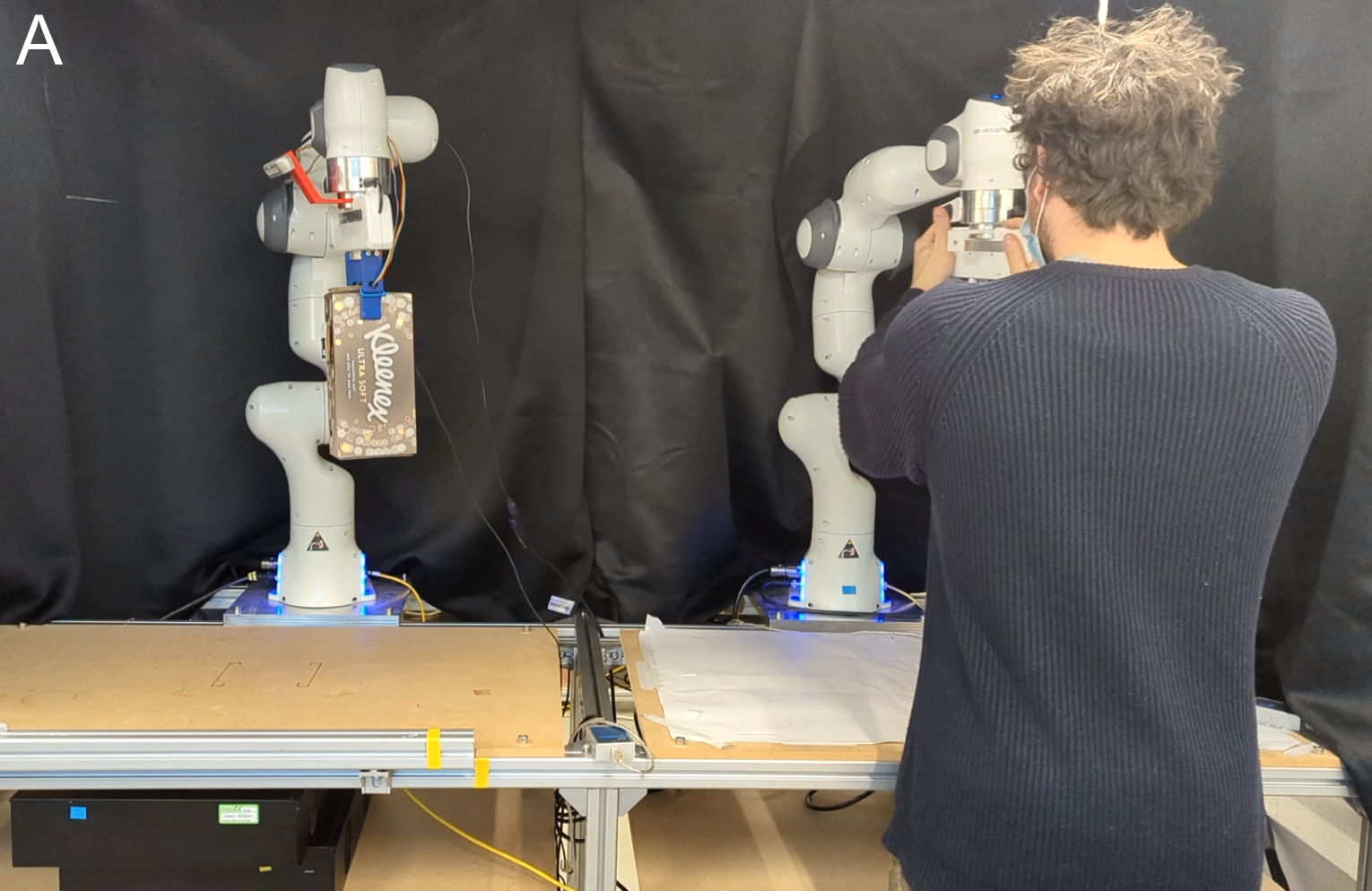}
                \includegraphics[trim=0pt 0pt 0pt 0pt, clip, height=0.23\textwidth]{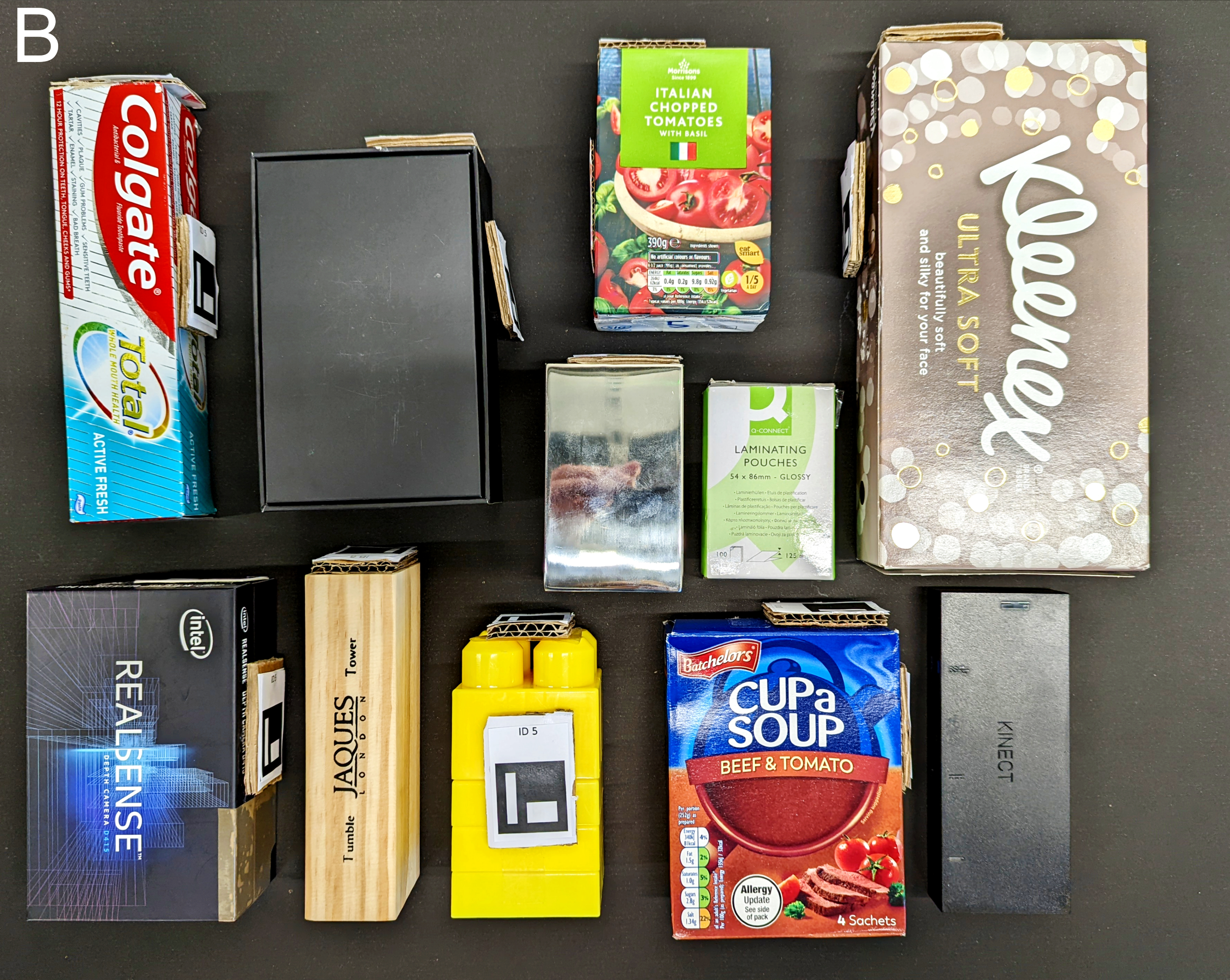}
                \includegraphics[trim=0pt 0pt 0pt 0pt, clip, height=0.23\textwidth]{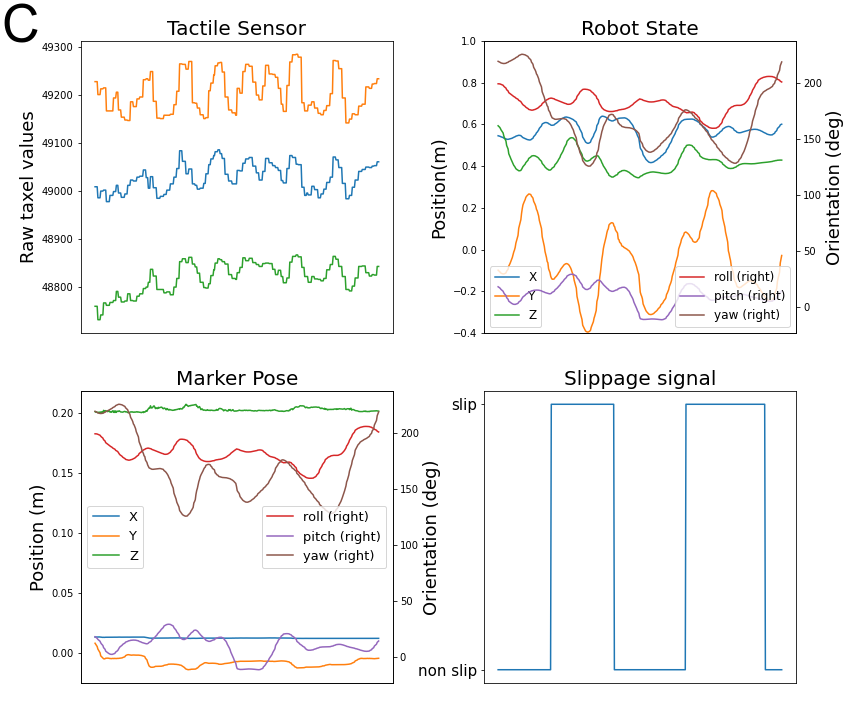}}
            \caption{(A) Teleoperated data collection set-up. The left robot is the `\emph{follower}', grasping the object with tactile sensing fingers. The right robot is the `\emph{leader}' and is teleoperated by human control (B) Eleven household box shaped objects used for training and testing, including tissue box, toothpaste box, and chopped tomatoes. Object set has variance in size, weight, centre of mass, material, stiffness and contact properties. Markers can be seen on the top and side of the objects, these are used to localise the object between the robot fingers, which is used for slip classification (C) Dataset example of a full trial.}
            \label{fig::Setup}
        \end{figure*}

    \paragraph{A novel dataset} containing highly dynamic and non-linear kinesthetic robot motions. Each trial contains a grasp and move motion, with a tactile enabled robot. The tactile sensors are low cost, low resolution (4x4 sensing elements of three forces) magnetic based sensors attached to both fingers of the robots pincer gripper\footnote{The uSkin sensor by xelarobotics.com}. We also track object position and orientation with respect to the gripper for slip labelling. The dataset contains 51,000 frames with 11 household objects, with 10.5\% of the data containing slippage cases. The dataset is publicly available \hyperlink{https://github.com/imanlab/action_conditioned_tactile_prediction}{here\footnote{https://github.com/imanlab/action\_conditioned\_tactile\_prediction}} 

    \paragraph{action-conditioned RNN} We develop two models -- one linear and one convolutional -- and show that they predict future fingertip tactile readings of a robot grasping different objects while moving through highly non-linear trajectories. We show that the models are capable of generalisation to unseen objects.

    \paragraph{Comparison study} We compare these models to state-of-the-art action-conditioned prediction models Convolutional Dynamic Neural Advection~\cite{finn2016unsupervised} (\textbf{CDNA}) used in \cite{tian2019manipulation}, Stochastic Video Predictor~\cite{babaeizadeh2017stochastic} (\textbf{SV2P}) and Stochastic Video Generator~\cite{denton2018stochastic} (\textbf{SVG}), as well as the non-action conditioned tactile prediction model PixelMotionNet~\cite{zhang2020towards} (\textbf{PMN}) and two basic multilayer perceptron (\textbf{MLP}) benchmark neural networks.

    The comparisons show that our proposed models outperform state-of-the-art approaches. We show this with (i) Quantitative analysis of test set prediction Mean absolute error (\textbf{MAE}), structural similarity (\textbf{SSIM}) and Peak Signal-to-Noise Ratio (\textbf{PSNR}) values, extended time horizon predictions and object generalisation, (ii) visual qualitative analysis of tactile prediction plots and (iii) predicted slip classification. \citet{nunes2020action} showed that analysis of time series prediction models should be done with respect to their use cases, so we selected slip prediction as a relevant use case for our predictive models.

\section{Related Works}
    \label{related_works}
    A large variety of tactile sensors have been developed in industry and literature, typically trading between resolution, affordability and sensitivity, \emph{image-based}\footnote{Such a technology includes a camera capturing deformation of a membrane.}~\cite{tian2019manipulation} and \emph{magnetic-based}~\cite{zhou2020learning} sensors. \emph{Video prediction} models have been applied to tactile image prediction using image-based tactile sensors~\cite{tian2019manipulation, zhang2020towards}. Finn et al.~\cite{finn2016unsupervised} introduced CDNA for video prediction. Tian et al.~\cite{tian2019manipulation} used the GelSight sensor~\cite{yuan2017gelsight} and proposed a deep tactile model predictive control system using CDNA to reach a goal tactile image in a simple task of object rolling. However, this work~\cite{tian2019manipulation} used small objects (dice and marbles) which could be completely contained within the sensors field of view. This suits the CDNA methodology, which uses convolutional kernels and object masks to move pixels about the input image to produce the next prediction. For larger objects this may not be the best approach as the system will need to predict force change not force motion and the object masks will not be able to locate any object. Studies reported the introduction of adversarial learning and learned priors, in SAVP~\cite{lee2018stochastic} and SV2P~\cite{babaeizadeh2017stochastic} respectively, improve on the CDNA architecture.

    \citet{denton2018stochastic} proposed a new SVG model that combines deterministic video prediction model, with time-dependent stochastic latent variables. The SVG architecture ``is competitive with other state-of-the-art video prediction models SAVP and SV2P'' \cite{villegas2019high}. Unlike SAVP and SV2P, the model is also made up entirely of standard neural network layers without any special computations like optical flow. Which, \citet{villegas2019high} argue, makes the model more generalisable.

    Magnetic-based sensors such as the Xela uSkin provide high frequency readings at each taxel with tri-axial readings. This sensor has several magnetic-based cells each measuring non-calibrated normal and shear forces, i.e. the readings are proportional to a normal and two shear forces. However, they provide low resolution when compared to vision based tactile sensors such as GelSight~\cite{yuan2017gelsight}, which comes at a cost of frequency and abstract readings. 
    Image-based tactile sensors benefit from the methods developed in computer vision~\cite{tian2019manipulation, finn2016unsupervised}. 
    Nonetheless, we chose to use the Xela uSkin magnetic based tactile sensor due to its low comparative cost, its high frequency readings which are essential for control and the extra challenge of analysing non-calibrated Xela readings (absolute value of the Xela sensor readings depends on the contact force and contact geometry). 
    \citet{zhou2020learning} converted the Xela uSkin tactile sensor readings to a visual representation that could be applied to the CDNA architecture. However, there are significant issues with the proposed representation. First, the resolution of the image reduces the resolution of the tactile readings; Second, the taxel objects cross over producing an impossible problem for the prediction model to interpret. Using this representation, \citet{zhou2020learning} proposed a simplified version of the CDNA model to perform tactile prediction. However, the CDNA inspired model produced poor test scores for the reasons outlined. \citet{zapata2018non, zapata2019learning} presented an image representation of the BioTac sensor from Syntouch and applied the ConvLSTM model for direction of slip classification. However, this model does not utilise robot actions.

    Tactile sensations are also used for improved grasping. \citet{zhang2020towards} proposed an improved grasping system through a new video prediction model called PixelMotionNet applied to the tactile images from FingerVision \cite{zhang2018fingervision}. However, these works only focus on grasp success rate by predicting contact and slip events while we focus on manipulation. Tactile based deep neural networks are also used for grasp policy learning~\cite{lee2019making}, slip detection~\cite{massalim2020deep}, tactile and visual data fusion for grasping~\cite{calandra2018more}, and tactile reinforcement learning for grasping~\cite{wu2019mat}.
    
    We apply our TPM's to a slip prediction/classification task. Slip detection methods are reviewed in \cite{chen2018tactile}. Heuristic approaches such as using a threshold on the rate of shear force change~\cite{kaboli2016tactile} or friction cone estimation~\cite{hogan2020tactile} are among the common methods. However, these methods cannot be generalized to novel objects or sensor types. Data driven classifiers such as SVM, Random Forest, CNNs, or RNNs showed good generalization results in slip detection to novel objects~\cite{veiga2020grip, meier2016tactile, deng2020grasping}. We apply Random Forest for slip classification in the tactile prediction space. 

    There are different datasets including tactile sensing. For instance, \citet{zapata2019tactile} used a household object dataset of 51 objects, recording more than 5500 grasps to test grasp stability using tactile sensation on novel objects. For slip classification, the authors used a second dataset of 11 objects. To predict and detect contact events with tactile sensation,
    ~\citet{zhang2020towards} generated a dataset of 11 items stating an ability to generalise across objects. For manipulation of a single object through tactile feedback alone, \citet{tian2019manipulation} created 3 datasets of 7400, 3000 and 4500 motions for three different objects. To perform a robust force estimation with image-based tactile sensors, \citet{sundaralingam2019robust} generated a dataset of 20,000 force samples and a dataset of 100,000 force samples (600 interactions).

    Video prediction models using LSTM recurrent layers have been applied to predict tactile data over time sequences. However, there is no comparison of such approaches. Model architecture, tactile data representation, use of non-conventional layers like optical flow or stochastic networks using learned priors have an unknown impact on the tactile prediction problem. In this work, we present two novel TPMs. We compare the performances of these models with those of state-of-the-art tactile prediction networks and sequence predictors via quantitative and qualitative studies as well as slip classification benchmark.

\section{Robot Manipulation Dataset}
    \label{dataset}
    One key real-world manipulation task is grasp and move motions. As the robot grasps and moves an object about its workspace, tactile sensations vary, depending on the object and the trajectory being performed. We collected a dataset with a range of human teleoperated kinesthetic motions, shown in Fig.~\ref{fig::Setup}-A, enabling a random and diverse dataset. The dataset consists of: (i) robot proprioception data in joint and task space, enabling action conditioning (ii) tactile data from both fingers of the gripper (iii) object position and orientation with respect to the robot's wrist. We used a human operator to intentionally create necessary accelerations in some cases to cause slippage for our case study in qualitative analysis. These motions are more complex than linear motions and more dense in slip cases and diverse tactile sensations than a random motion based dataset.

    To ensure the dataset is realistic to real-world scenes, we use a set of common flat-surfaced household objects, shown in Fig.~\ref{fig::Setup}-B. We collected two datasets: a train dataset, consisting of 52 trials (40,000 frames) with 9 objects; and a test dataset, of 22 trails (11,000 frames) with 3 objects, two of which are not present in the train dataset. Examples of the dataset trails are shown in Fig.~\ref{fig::Setup}-C. The dataset was collected at 40 frames per second, the maximum frame rate of the tactile sensors. The Xela uSkin tactile sensor contains 16 sensing elements arranged in a square grid, each outputting shear x, shear y and normal forces (Fig.\ref{fig::intro_image}-B). The Xela sensors high frequency enables more aggressive and fast robot motions that can create the object slippage and larger tactile changes which we require for our dataset.

    The constraint on flat surfaces for grasping produces a more consistent task across trials for the models to capture when compared to objects with varied grasp surface topology. The low resolution (4.7 [mm] distance between two sensing points) tactile sensor may also struggle to capture more complex surfaces. This flat-surfaced objects dataset presents a baseline of objects and movements useful for tactile prediction tasks, we leave progression to more complex surface typologies as future work.

    To observe the state of the object with respect to the gripper, and to enable slip classification during the trials we recorded the objects pose ($SE(3)= \mathbb{R}^3 \ltimes
    SO(3)$ where $SO(3)$ is a group of rotation in 3-D space expressed by Euler angles, i.e. $\in \mathbb{R}^{3}$) using a wrist camera and ArUco markers \cite{garrido2014automatic} on the top and one side of each object. Using two markers ensures at least one marker is in camera field of view providing a continuous recording of position and orientation of the object. Inspired by \citet{begalinova2020self}, we applied Cumulative Sum anomaly detection \citet{hinkley1971inference} on the Z component of the object position in robot wrist frame to classify slip as a binary signal.

    \begin{figure}[tb!]
        \centering

        \includegraphics[height=0.235\textwidth, frame]{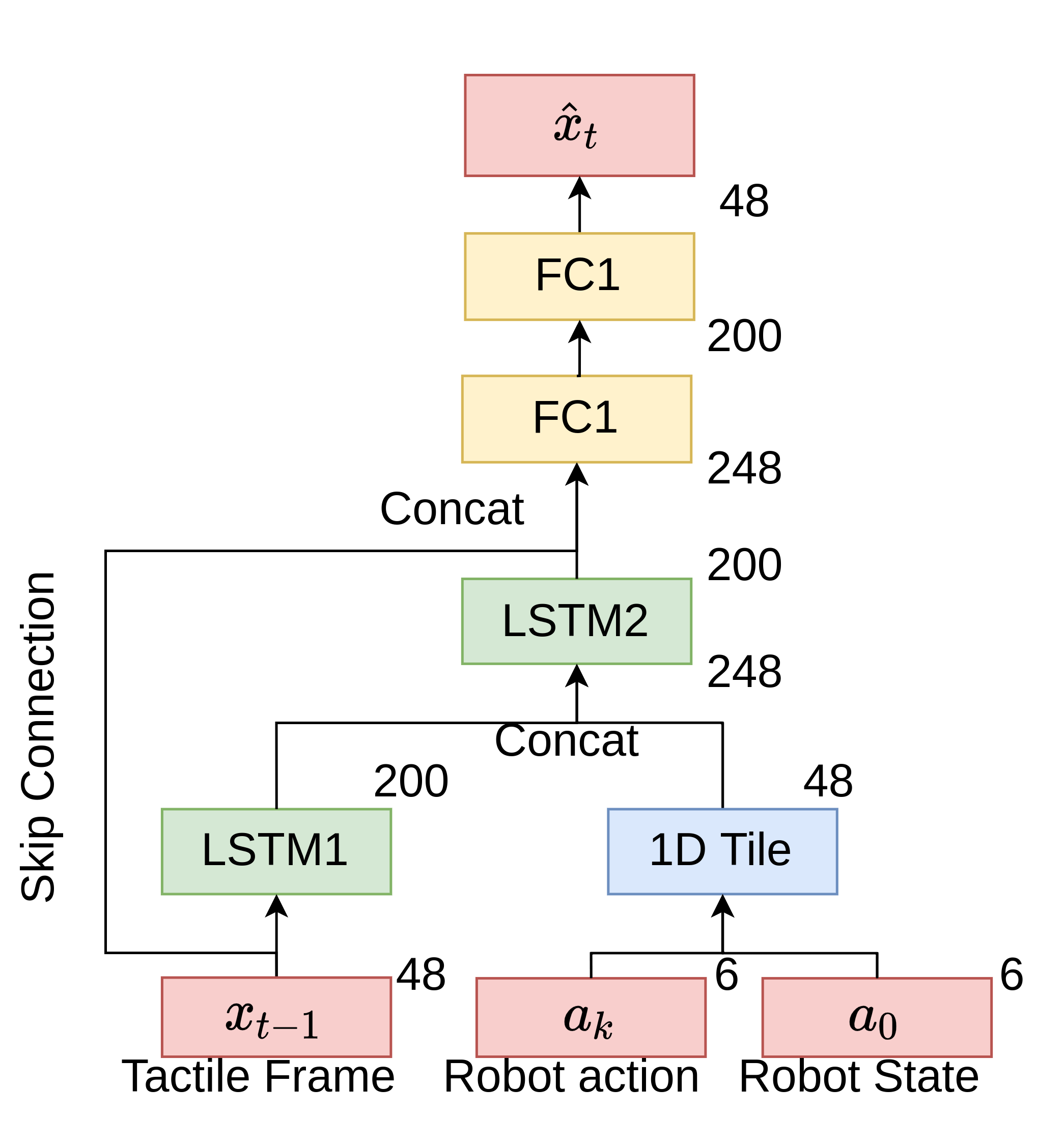}
        \includegraphics[height=0.235\textwidth, frame]{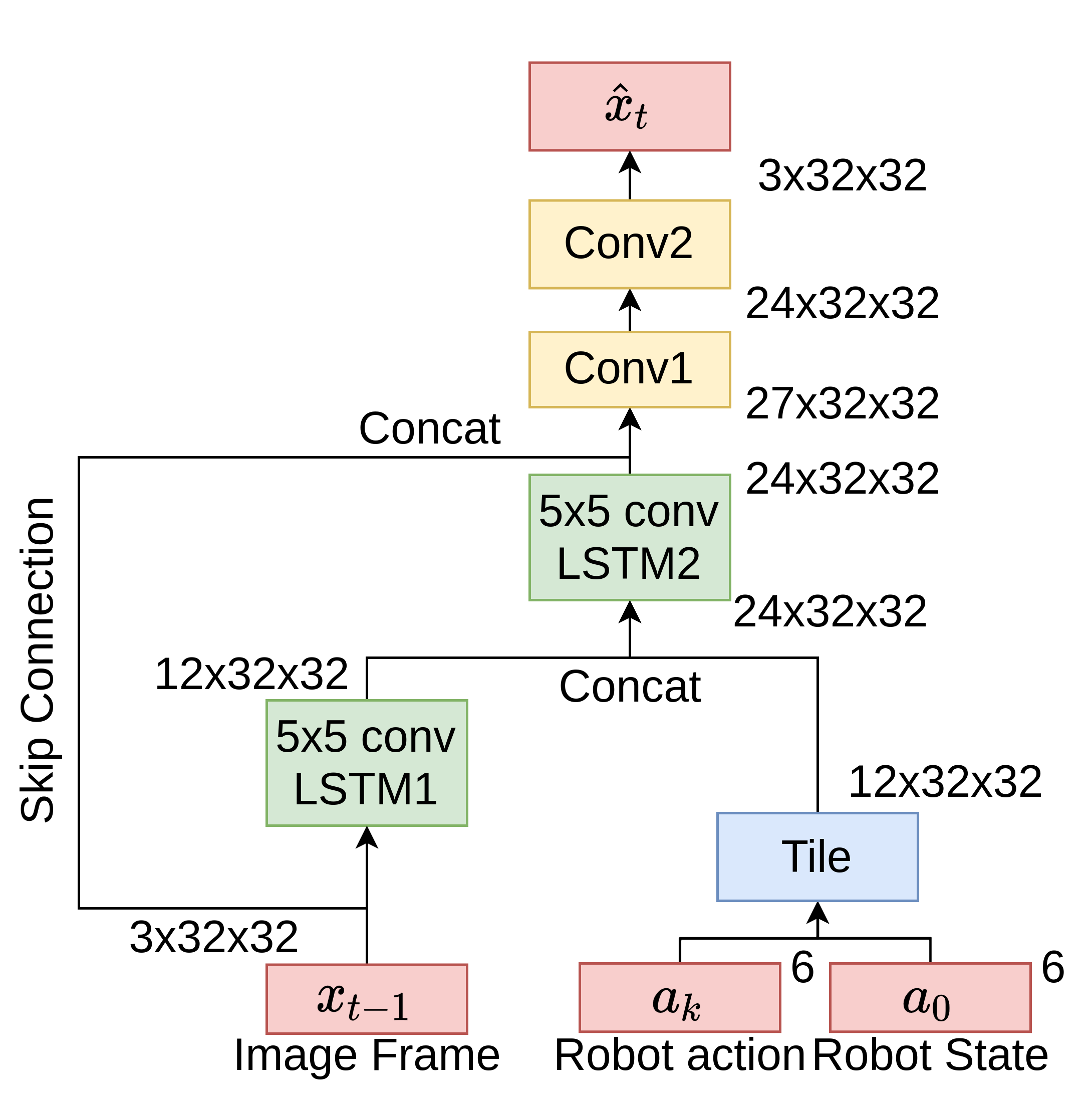}
        \caption{Tactile prediction model architectures (left) Action Conditioned Tactile Prediction (ACTP) and (right) Action Conditioned Tactile-Video Prediction (ACTVP)}
        \label{fig::ACTPandACTVP}
    \end{figure}

\section{Action Conditioned Tactile Predictive Models}
    One of the major objectives of this work is to use state-of-the-art data-driven predictive models and adapt/utilise them for predicting tactile sensation during physical robot interactions. We compare and analyse the performance of these models in predicting tactile signals of a magnetic-based sensor, which are contact geometry/properties dependent in real-world manipulation tasks, making calibrating such sensors challenging. Our main assumptions include (1) models will have access to the future/planned robot states as well as (2) the previous tactile readings during the trial. The models should predict for a future time horizon. These assumptions are useful and in line with requirements of many control strategies, e.g. model predictive control.

    Our developed models perform conditional predictions based on a set of \textit{c context} frames $\textbf{x}_{0},...,\textbf{x}_{c-1}$. These context frames are previous readings from the interaction. Our target is to sample from $p(\textbf{x}_{c:T}|\textbf{x}_{0:c-1})$ where $\textbf{x}_{i}$ denotes the i$^{th}$ tactile frame in the sequence and $T$ is the sum of the context frame length and the prediction horizon length.

    Our problem of action-conditioned tactile prediction can be defined as, a model must predict a sequence of future tactile states $\textbf{x}_{c:T}$ given a sequence of previous robot actions $\textbf{a}_{0:c-1}$, previous tactile states $\textbf{x}_{0:c-1}$ and a sequence of future/planned robot actions/trajectory $\textbf{a}_{c:T}$. A robot action, $\textbf{a}\in\mathbb{R}^{6}$, is the end-effector task space position and orientation (Euler angles) with respect to the robot base, while a tactile sample is $\textbf{x} \in\mathbb{R}^{16 \times 3}$.
    
            \begin{equation}
                p(\textbf{x}_{c:T}|\textbf{x}_{0:c-1}, \textbf{a}_{0:T})   
            \end{equation}

    Factorising this we can define the model as $\Pi^{T}_{t=c}p_{\theta}(\textbf{x}_{t}|\textbf{x}_{0:t-1}, \textbf{a}_{0:t})$. Learning now involves training the parameters of the factors $\theta$.
    
    We create two bespoke model for the tactile prediction task (Fig.~\ref{fig::ACTPandACTVP}): (i) Action Conditioned Tactile Prediction network (\textbf{ACTP}) and (ii) Action Conditioned Tactile Video Prediction network (\textbf{ACTVP}). The two models define the difference in potential representation of the tactile date. First, the tactile data can be flattened from $\textbf{x} \in \mathbb{R}^{16 \times 3}$ to $\textbf{x} \in \mathbb{R}^{48}$ features, and used in a linear network. Second the data can be scaled up to an image of $\textbf{x} \in \mathbb{R}^{32 \times 32 \times 3}$ which enables the application of convolutional layers and convLSTMs. We keep the structure of the two models the same outside of this, enabling a more direct comparison to tactile data representation.

    The model structure uses tiling to upscale the robot states to the same shape as the tactile data. The model takes inspiration from the current state-of-the-art tactile prediction network `PixelMotionNet'~\cite{zhang2020towards}, using two LSTMs then two linear layers. However, we take equal inspiration from the concatenation process of CDNA~\cite{finn2016unsupervised} which concatenates the robot state and action data in the middle of the LSTM chain. We also use skip connections in the same manner. We do not apply the optical flow approaches shown in PMN~\cite{zhang2020towards} and CDNA. This enables comparison between the optical flow method PMN and our models. 

    In time-steps $t$, our action conditioned models sequence through $\{t-c: t\}$. Once all the context data (i.e. previous robot and tactile states) have been fed to the model, it then predicts the future tactile frames $\hat{\textbf{x}}$ from time-step $t+1$ to $t+T$, where the predicted tactile frame at time step $i$ becomes the input to the model for the next time-step $i+1$.
    
    \citet{villegas2019high} showed specialised, handcrafted architectures reduce a model's generalisation to new applications and that simple model layers are beneficial for this reason. As such, we use PMN's simple architecture as the key motivation. Likewise, the motivation for the prediction architecture was also based on the areas left by existing models: (i) the two models presented have the exact same underlying shape to enable tactile data representation comparison (ii) optical flow was explored in PMN and CDNA, in different ways, so we did not include it in the presented models. (iii) The position of action data concatenation is in the middle of the LSTM chain and differs from action-conditioned PMN, which integrates at the beginning.


\section{Results and discussion}

    To evaluate the performances of the proposed predictive models for tactile signal prediction, we performed three different studies (1) quantitative and (2) qualitative comparisons of the tactile predictions and (3) a slip prediction benchmark.
    
    We aim to identify the best performing models and key features for tactile prediction during robot manipulation. Table \ref{tab:model_features} shows each model tested with its key features. To explore the features of PMN~\cite{yamaguchi2016combining}, we adjust the original architecture to create (1) \textbf{PMN-AC}, an action conditioned version of PMN and (2) \textbf{PMN-AC-NA}, the PMN-AC model without the final addition stage, this enables us to explore the effect of action conditioning and optical flow. We also include CDNA~\cite{finn2016unsupervised}, which is a much larger optical flow based model and has been proven to work for tactile prediction of vision based tactile sensors in \cite{tian2019manipulation}. ACTP and ACTVP have the same structure, however, ACTP uses the raw tactile values, where as ACTVP uses an image representation of the tactile data, enabling exploration of the positives and negatives of converting to image representations. Moreover, we explore state of the art video prediction model SVG~\cite{denton2018stochastic}, with 4 LSTM layers for prediction pipeline and 3 LSTM layers for the prior. This comparison shows the current state-of-the-art video prediction performance on this dataset. Furthermore, this helps us to discuss the use of learned priors and encoder decoder models in this setting. Finally, we include two simple baseline MLP models as simplified benchmarks, \textbf{MLP} and \textbf{MLP-AC} (which is identical but with action conditioning).

    We chose to test the models with a prediction horizon of 10 time steps (0.25 seconds in real time). Although the cut off point is arbitrary in our current setting, we chose this point due to high speed of motion in comparison with similar video prediction works where prediction can be pushed to 1 second. Our methods can be easily adapted for longer prediction horizon.

    \begin{table}[t]
        \centering
        \caption{Key features of tested models; Action-conditioned (AC), Optical Flow (OF); stochastic model (St); Image and linear based tactile data representation (Image \& Linear); Encoder-Decoder (ED).}
        \label{tab:model_features}
        \begin{tabular}{|c|c|c|c|c|c|c|}
            \hline
            Model & AC & OF & St & Image & Linear & E-D\\\hline
            PMN-AC    & \checkmark  & \checkmark    & $\times$ & \checkmark & $\times$ & $\times$  \\
            PMN-AC-NA & \checkmark  &  $\times$       &  $\times$ & \checkmark & $\times$  & $\times$  \\
            PMN       &  $\times$     & \checkmark    & $\times$  & \checkmark &  $\times$ &  $\times$ \\
            ACTP      & \checkmark  &$\times$         &$\times$   &  $\times$ & \checkmark &  $\times$ \\
            ACTVP     & \checkmark  &  $\times$       & $\times$  & \checkmark &  $\times$ &  $\times$ \\
            MLP       &  $\times$     & $\times$        &  $\times$ &  $\times$ & \checkmark &  $\times$ \\
            MLP-AC    & \checkmark  &  $\times$       & $\times$  & $\times$  & \checkmark &  $\times$ \\
        CDNA      & \checkmark  & \checkmark        &  $\times$ & \checkmark & $\times$  &  $\times$ \\
            SVG       & \checkmark  &  $\times$ & \checkmark & \checkmark & $\times$  & \checkmark \\
            \hline
        \end{tabular}
    \end{table}

    \begin{table*}[t]
        \centering
        \caption{Model performance per object and generalisation accuracy}
        \label{tab:train_vs_test}
        \resizebox{\textwidth}{!}{
        \begin{tabular}{|c||c|c|c|c|c|c|c|c|c||c|c|c||c|c|c|}
            \hline
            & \multicolumn{9}{|c||}{Training Objects MAE $\times$ 100}&\multicolumn{3}{|c||}{Test Objects MAE $\times$ 100 }&\multicolumn{3}{|c|}{Overall MAE $\times$ 100}\\
            \hline
            Model & \shortstack{Tooth \\ paste} & \shortstack{Metal \\ Cube} & \shortstack{Lego} & \shortstack{Ink \\ Box} & \shortstack{Soup \\ Box} & \shortstack{Toma- \\toes} & \shortstack{Wood \\ Block} & \shortstack{Tissue \\ Box} & \shortstack{Matt \\ Box} & \shortstack{Wood \\ block} & \shortstack{Power \\ Unit} & \shortstack{Intel \\ Box} & Seen & Novel & Dif \\\hline
            PMN-AC & 0.770 & 0.627 & \textbf{0.561} & 0.428 & 0.360 & 1.127 & 0.625 & 0.890 & 0.719 & 0.948 & 0.683 & 0.715 & 0.948 & 0.704 & 0.244\\
            PMN-AC-NA & \textbf{0.746} & \textbf{0.584} & 0.569 & \textbf{0.416} & \textbf{0.348} & \textbf{1.044} & \textbf{0.582} & \textbf{0.786} & 0.706 & 0.910 & 0.594 & 0.649 & 0.845 & 0.629 & 0.215\\
            PMN & 0.919 & 0.811 & 0.671 & 0.511 & 0.443 & 1.294 & 0.704 & 0.943 & 0.741 & 1.060 & 0.739 & 0.810 & 0.987 & 0.784 & 0.203\\
            ACTP & 1.580 & 1.816 & 1.270 & 0.972 & 0.897 & 3.029 & 1.281 & 1.664 & 1.273 & 2.253 & 1.866 & 1.870 & 2.098 & 1.861 & 0.236\\
            ACTVP & 0.799 & 0.596 & 0.576 & 0.427 & 0.353 & 1.131 & 0.606 & 0.905 & \textbf{0.641} & \textbf{0.736} & \textbf{0.579} & \textbf{0.612} & \textbf{0.693} & \textbf{0.598} & \textbf{0.095}\\
            MLP & 1.548 & 1.717 & 1.524 & 1.184 & 1.140 & 2.164 & 1.432 & 1.712 & 1.594 & 1.767 & 1.429 & 1.447 & 1.767 & 1.441 & 0.326\\
            MLP-AC & 1.747 & 1.910 & 1.736 & 1.406 & 1.352 & 2.342 & 1.580 & 1.882 & 1.767 & 1.917 & 1.606 & 1.656 & 1.917 & 1.639 & 0.278\\
            CDNA & 18.988 & 16.614 & 17.448 & 19.414 & 21.872 & 17.038 & 17.610 & 21.041 & 17.497 & 37.88 & 40.10 & 41.28 & 37.88 & 40.69 & -2.809\\
            SVG & 3.942 & 7.328 & 4.508 & 2.829 & 2.986 & 6.435 & 4.172 & 3.430 & 5.670 & 4.894 & 5.385 & 3.531 & 4.619 & 4.458 & 0.161\\\hline
        \end{tabular}}
    \end{table*}


    \paragraph{Quantitative Comparison:} 

        \begin{table}[h]
            \centering
            \caption{Model Performance (entire event horizon) on grasp and move test dataset. Mean absolute error (MAE), Structural Similarity (SSIM) and Peak Signal-to-Noise Ratio (PSNR)}
            \label{tab:full_perf_table}
            \begin{tabular}{|c|c|c|c|}
                \hline
                Model & MAE & PSNR & SSIM \\\hline
                PMN-AC & 0.00782 & 91.8009 & 0.9956 \\
                PMN-AC-NA & 0.00720 & 91.5760 & 0.9956 \\
                PMN & 0.00854 & 89.5510 & 0.9910 \\
                ACTP & 0.01943 & - & - \\
                ACTVP & \textbf{0.00631} & \textbf{91.8266} & \textbf{0.9965} \\
                MLP & 0.01545 & - & - \\  
                MLP-AC & 0.01727 & - & - \\  
                CDNA & 0.39879 & 49.0997 & 0.7219 \\
                SVG & 0.0455 & 74.6662 & 0.8656 \\
                \hline
            \end{tabular}
        \end{table}

        Table \ref{tab:full_perf_table} shows a comparison of MAE, PSNR and SSIM across the tactile prediction models. We use these three metrics as MAE gives a basic quantitative comparison across image and non-image based models. To make comparison between the video prediction models we show average standard metrics PSNR \cite{huynh2008scope} and SSIM \cite{wang2004image}. SSIM shows the similarity between two images and is used for basic comparison, PSNR penalises outlier values so indicates models that produce these.

        Our ACTVP tactile prediction model has the best performance across these metrics, outperforming the state of the art prediction model SVG and CDNA. This suggest that the image representation of tactile values, which encodes the topology of the sensor values, has a beneficial impact over the flattened 48 feature vector used by ACTP as the models are identical outside of this change.

        Comparing PMN with PMN-AC, we observe that action conditioning has a positive response with respect to the performance metrics. Changes in tactile data are created due to changes in robot action so this finding was to be expected.

        Observing differences between PMN-AC and PMN-AC-NA, we observe the optical flow approach has a negative impact on performance with respect to the predicted 48 tactile features, despite producing better image quality with reduced outlying errors. This could be due to the approach of optical flow methods, which calculate changes to the previous image as supposed to creating the next through the network, which emphasises the values of the previous image.

        SVG and CDNA produce poor prediction results across all performance metrics. SVG uses an encoder decoder structure to represent the tactile features, this result indicates that encoding the tactile features has a negative impact on prediction capability in this scene. CDNA generates the worst prediction accuracy, likely due the assumptions made by CDNA (stated section~\ref{related_works}) which not applicable in our dataset, SVG does not make these assumptions and so produces stronger performance.

        The dataset contains \emph{seen} and \emph{unseen} objects. This helps to analyse the generalisation ability of models across the dataset (see table \ref{tab:train_vs_test}). We observe that models are capable of maintaining prediction accuracy when generalising to new objects. Table \ref{tab:ts_perf_table} also suggest th prediction accuracy over time is decreasing. ACTVP has the lowest t+10 prediction error (outside of the poor performance models), suggesting the model preforms best at time series prediction of the tactile data during manipulation tasks. Equally, we see that removing optical flow measures from a prediction model of PMN results in increased performance over extended time horizons. From quantitative analysis we conclude that the best performing model is ACTVP. While we find that action conditioning of prediction models is beneficial, learned priors and optical flow techniques have no observable benefit.

        \begin{table*}[t]
            \centering
            \caption{Model performance for prediction time steps t+1, t+5, t+10}
            \label{tab:ts_perf_table}
            \begin{tabular}{|c||c|c|c|c||c|c|c||c|c|c|}
                \hline
                & \multicolumn{4}{|c||}{MAE}&\multicolumn{3}{|c||}{PSNR}&\multicolumn{3}{|c|}{SSIM}\\
                \hline
                Model & t+1 & t+5 & t+10 & t+1 - t10 & t+1 & t+5 & t+10 & t+1 & t+5 & t+10 \\\hline
                PMN-AC & \textbf{0.00156} & 0.00732 & 0.01358 & 0.0120 & \textbf{106.1998} & 93.0509 & 87.6827 & \textbf{0.9991} & 0.9892 & 0.9750 \\
                PMN-AC-NA & 0.00370 & 0.00609 & 0.0120 & 0.008 & 95.9277 & 93.7079 & \textbf{89.2150} & 0.9969 & 0.9912 & \textbf{0.9780} \\
                PMN & 0.00449 & 0.00768 & 0.01352 & 0.0090 & 94.1486 & 91.0600 & 86.9993 & 0.9885 & 0.9876 & 0.9736 \\
                ACTP & 0.01395 & 0.01878 & 0.02484 & 0.0109 & - & - & - & - & - & - \\
                ACTVP & 0.00302 & \textbf{0.00543} & \textbf{0.01128} & 0.0083 & 96.7655 & \textbf{94.2063} & 89.0712 & 0.9983 & \textbf{0.9913} & 0.9778 \\
                MLP & 0.01314 & 0.01345 & 0.01943 & \textbf{0.0063} & - & - & - & - & - & - \\  
                MLP-AC & 0.01227 & 0.01623 & 0.02089 & 0.0086 & - & - & - & - & - & - \\  
                CDNA & 0.12585 & 0.20829 & 1.36854 & 1.24269 & 64.7118 & 60.1571 & 41.5055 & 0.9032 & 0.808 & 0.3964 \\
                SVG & 0.03419 & 0.04584 & 0.05455 & 0.0204 & 78.2405 & 75.0495 & 73.2830 & 0.8811 & 0.8661 & 0.8494 \\\hline
            \end{tabular}
        \end{table*}


    \paragraph{Qualitative analysis}
        We can better understand the performance of tactile prediction models through visual examination. In this section, we highlight some of the most important visual differences between the tactile prediction models. We present tactile predictions on the last context frames time-step, replicating the setup of a control scenario. One key visual feature we use to judge model performance is the time step that a model predicts a peak or trough. Models that show these changes in tactile force indicate prediction of tactile sensation, the earlier they're predicted, the better. First, we observe across all models an inability to predict during the initial object grasping phase, this is due to not providing the robots finger states to the models as well as there being no prior knowledge about the object being grasped or the position of grasp on that object.

\begin{figure}[tb!]
            \centering
            \adjustbox{fbox= 0pt 1pt, frame}{\includegraphics[width=0.47\textwidth]{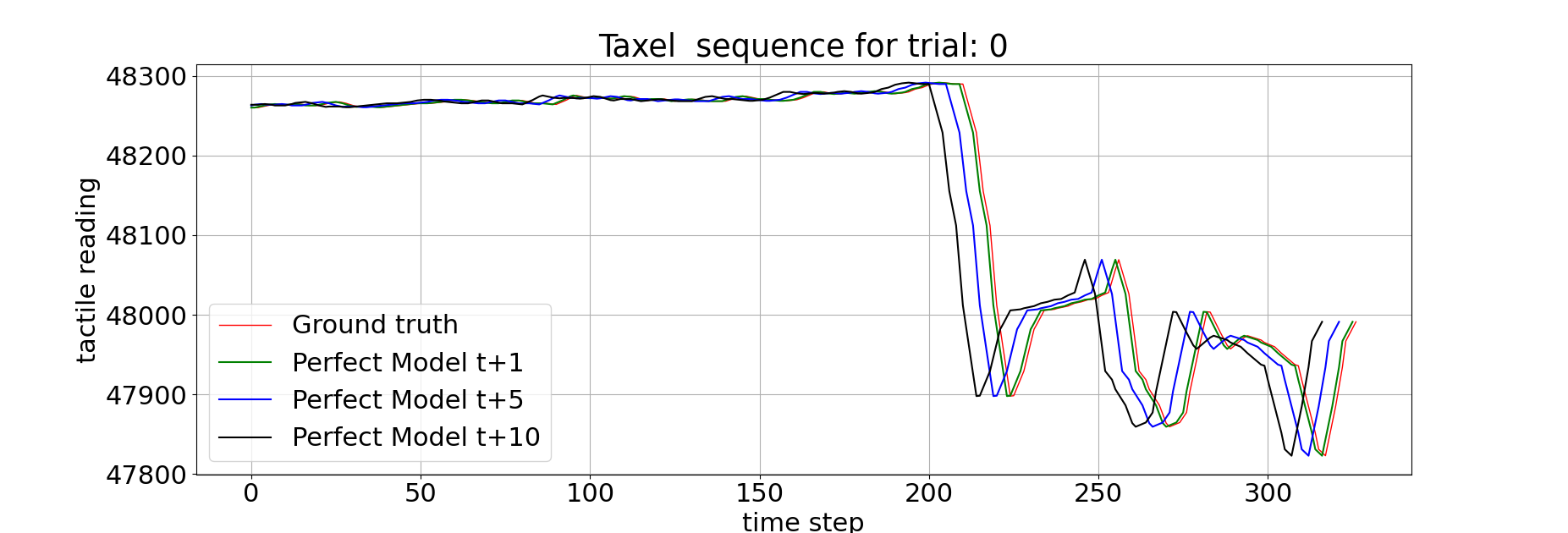}}

            \vspace{0.2cm}

            \adjustbox{fbox= 0pt 1pt, frame}{\includegraphics[width=0.47\textwidth]{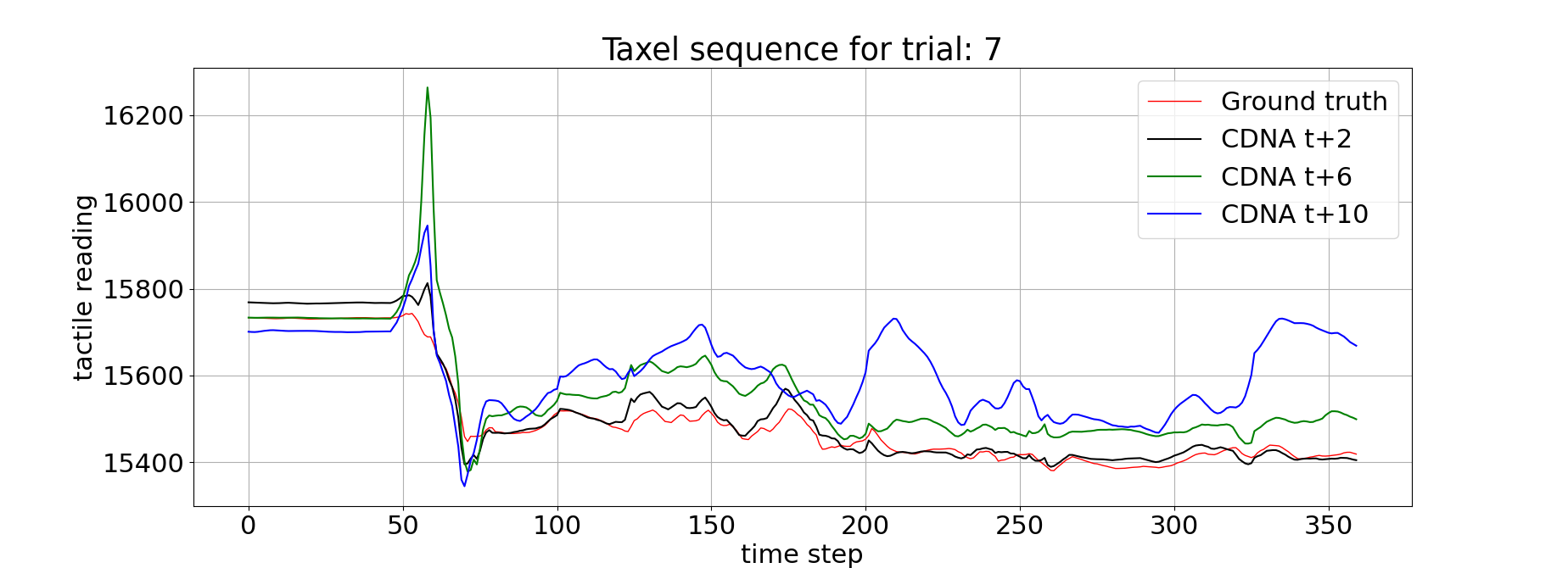}}
            \caption{Tactile predictions at prediction time-step (top) examples the perfect tactile prediction model for reference with Figures \ref{fig::intro_image}, \ref{fig::t10plotsImages}, \ref{fig::ACcomparison}, \ref{fig::NAvsA}, \ref{fig::ACTPvsACTVP} and \ref{fig::classifier}. (bottom) CDNA predictions, showing poor performance, especially at extended time horizons.}
            \label{fig::PerfectVsCDNA}
        \end{figure}

        For comparison with other prediction plots, we show a 'perfect' model's predictions in Fig.~\ref{fig::PerfectVsCDNA}. CDNA's poor performance metrics correlate with equally poor tactile signal predictions, Fig.~\ref{fig::PerfectVsCDNA}, shows the rapid degradation of the tactile predictions over the prediction horizon. Despite better performance metrics than CDNA and SVG, the two MLP benchmark models produce noisy representations of a naive system, replicating the last context frame for the entire prediction horizon.

\begin{figure}[tb!]
            \centering
            \adjustbox{fbox= 0pt 1pt, frame}{\includegraphics[width=0.47\textwidth]{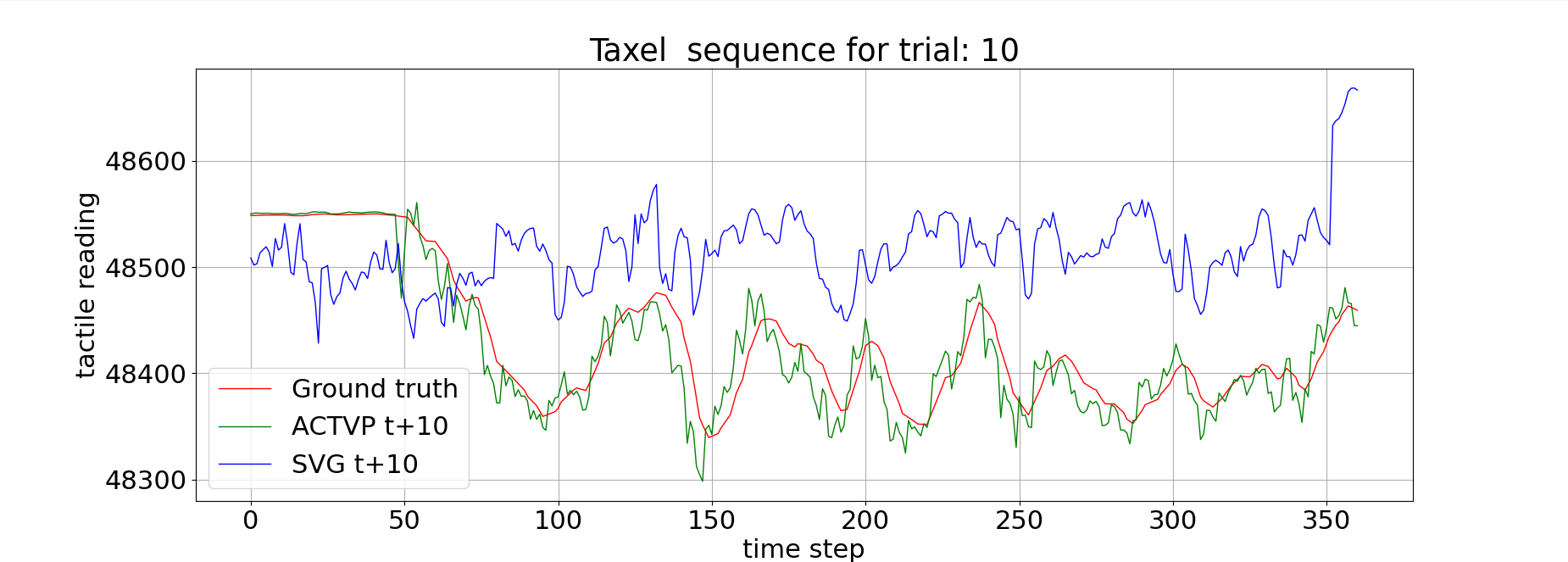}}
            \caption{Comparison between SVG and ACTVP, showing the poor performance of SVG's t+10 predictions, this level of performance is also indicated by the performance metric results shown in Table \ref{tab:full_perf_table}.}
            \label{fig::t10plotsImages}
        \end{figure}

        Fig.~\ref{fig::t10plotsImages} shows a comparison between SVG and ACTVP. The SVG predictions are significantly worse with respect to the ACTVP, this is mirrored by SVG's poor performance metrics too. Opposite to this, ACTVP's predictions attempt to predict change in taxel values, despite being noisy. Kalman filtering could be used to reduce this noise, however, it would have a detrimental effect on reducing the time difference shown between the predicted changes in tactile data.

\begin{figure}[tb!]
            \centering
            \adjustbox{fbox= 0pt 1pt, frame}{\includegraphics[width=0.47\textwidth]{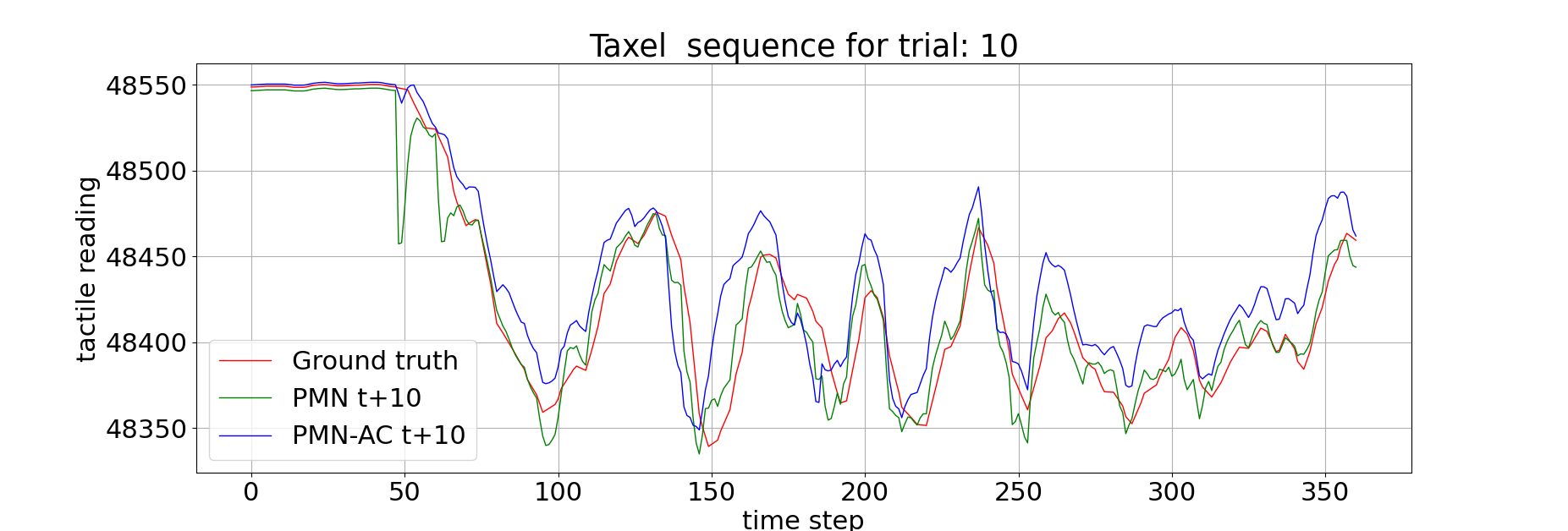}}
            \caption{Comparison between action conditioned and non action conditioned PixelMotionNet, showing similar performance on their t+10 predictions. The peaks and troughs of tactile prediction models are shown prior to the ground-truth tactile signals changes suggesting ability to predict tactile data.}
            \label{fig::ACcomparison}
        \end{figure}

\begin{figure}[tb!]
            \centering
            \adjustbox{fbox= 0pt 1pt, frame}{\includegraphics[width=0.47\textwidth]{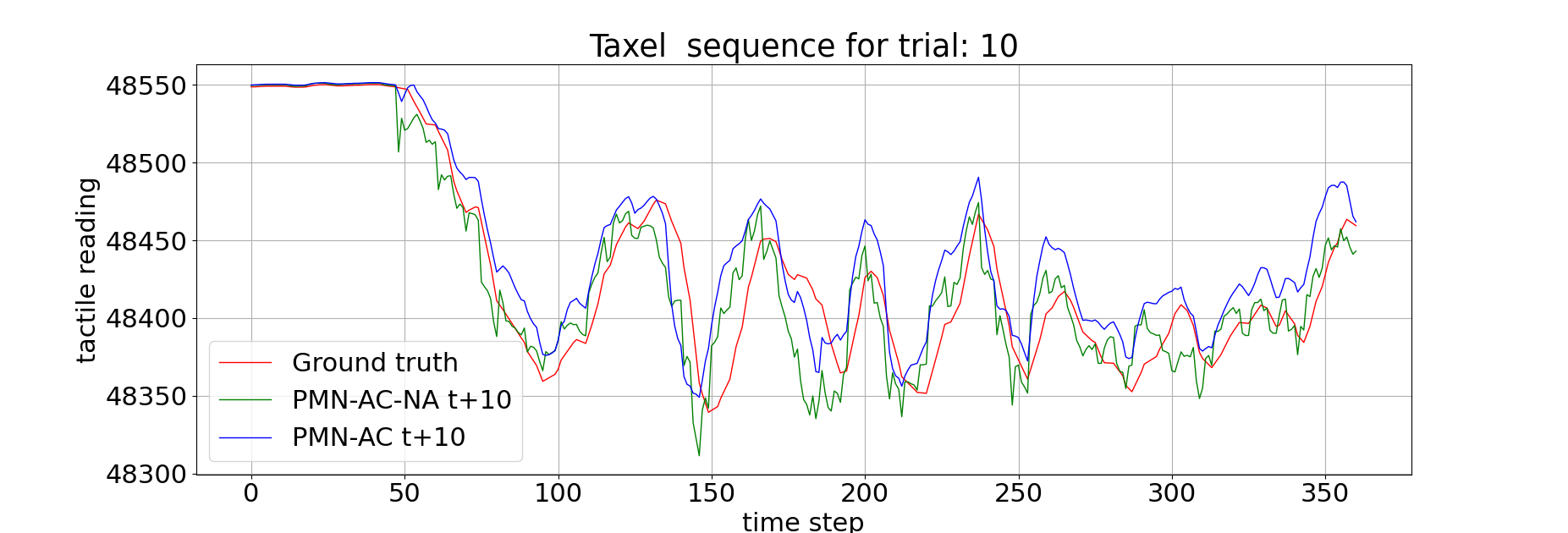}}
            \caption{Comparison between action conditioned PixelMotionNet with and without the optical flow addition layer.}
            \label{fig::NAvsA}
        \end{figure}

        Fig.~\ref{fig::ACcomparison} shows PMN and PMN-AC models produce similar predictions, suggesting low impact from action conditioning on tactile prediction performance. Fig.~\ref{fig::NAvsA} indicates that the inclusion of the optical flow layer in PMN, results in tactile predictions closer to the true values. Although the optical flow based PMN-AC appear to produce smoother predictions, we do not see indication of improved tactile prediction with respect to peaks and troughs.

\begin{figure}[tb!]
            \centering
            \adjustbox{fbox= 0pt 1pt, frame}{\includegraphics[width=0.47\textwidth]{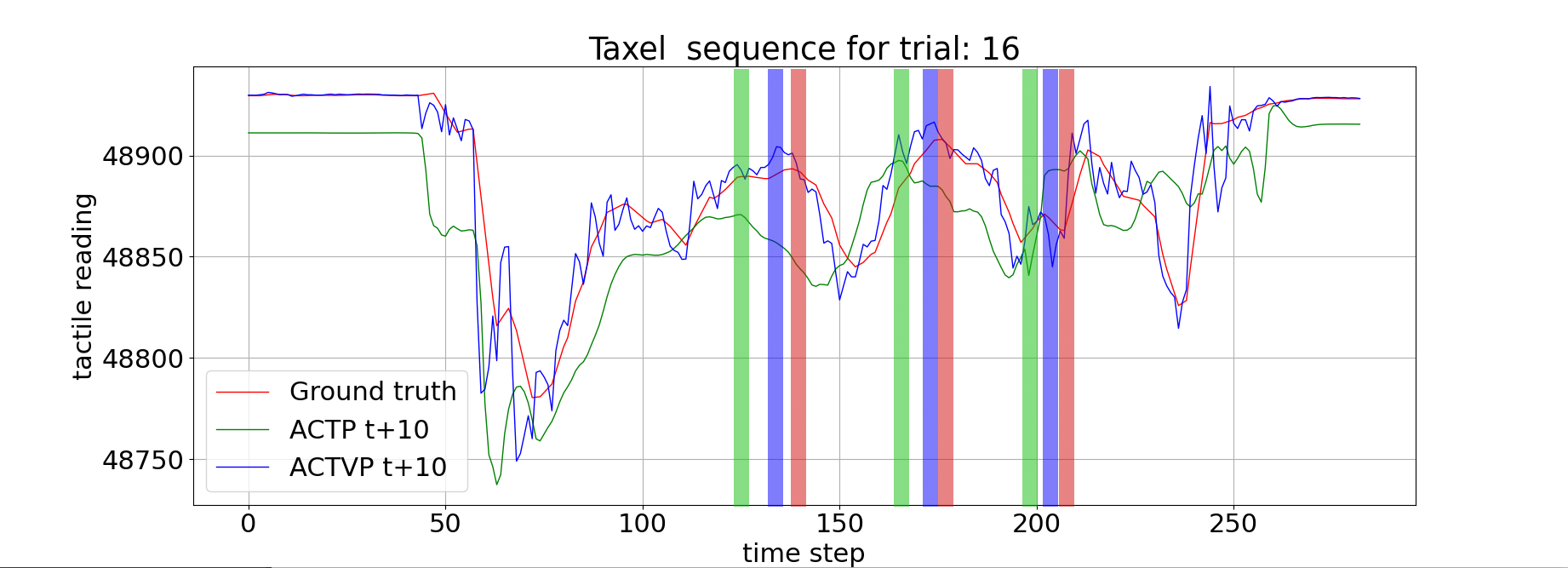}}
            \caption{Comparison between ACTP and ACTVP, showing ACTP predictions are ahead of ACTVP's, however with significant offset in taxel value.}
            \label{fig::ACTPvsACTVP}
        \end{figure}

        Comparing the two novel models, shown in Fig.~\ref{fig::ACTPvsACTVP}, we can highlight two shortcomings of relying only on the performance metrics. The peaks and troughs of the prediction models and the ground-truth signal are shown in highlighted bars. The ACTP predictions are worse than ACTVP with respect to the taxel values (Y-axis), which is indicated in the performance metrics. However, ACTP's predictions of changes in taxel values, indicated by peak and trough points, are shown to be significantly better than ACTVP. Second we can also observe a far smoother prediction plot with ACTP. This suggests that scaled up image representations of the tactile data have a negative impact on prediction performance. Overall, we find that performance metrics and even the loss functions used to train these systems may not fully indicate strong model performance. We conclude that despite poor performance metrics, visual assessment indicates that ACTP is the best performing model overall, followed by ACTVP.

\paragraph{Slip Classification}
We use Random Forest for slip classification~\cite{biau2016random} of the predicted tactile. In a slip prediction study, \cite{veiga2015stabilizing} shows Random Forest outperforms other classification approaches. We trained separate Random Forest classifiers on each model and a classifier on the raw tactile data as well. F1 score is metric suitable for comparing classification performance. Nonetheless, in our slip prediction setting, we are concerned with how often the prediction system predicts slip in prediction horizon before it actually occurs. We use two extra metrics \emph{score 1} and \emph{score 2} to measure the performance of classifier.

    \begin{equation}
        Score = (C1 \times f1) + (C2 \times s1) + (C3 \times s2)
    \end{equation}
    Where:
    \begin{align*}
        f1 &= 2 * \frac{\textrm{precision}*\textrm{recall}}{\textrm{precision} + \textrm{recall}} \\
        s1 &= \frac{\textrm{slip predictions}}{\textrm{num slip instances}} \\
        s2 &= 0.1 * \textrm{prediction to detection distance}
    \end{align*}

Where, f1, s1, and s2 account for detection rate, prediction rate, and prediction horizon, respectively. $C1=1$, $C2=0.1$, and $C3=0.2$ are coefficients indicating each terms influence on the score. While f1 is the most important term, s2 is given a slightly larger coefficient than s1 since the prediction horizon is considered more important than the prediction frequency. 

Performing evaluation with a real world application of tactile prediction provides a more realistic understanding of model performance when compared to the previous performance metrics. ACTP and PMN\_AC have the highest prediction scores and PMN and CDNA the lowest ones. Since the customised score holds both the detection and prediction of slippage it can be stated that models with higher scores show overall better detection and prediction behaviour. It can be observed that action-conditioning PMN improved its slip prediction performance.

    \begin{figure}[tp]
        \vspace{-20pt}
        \adjustbox{fbox= 0pt 1pt, frame}{\includegraphics[trim=0 0 0 0, clip, width=0.99\columnwidth]{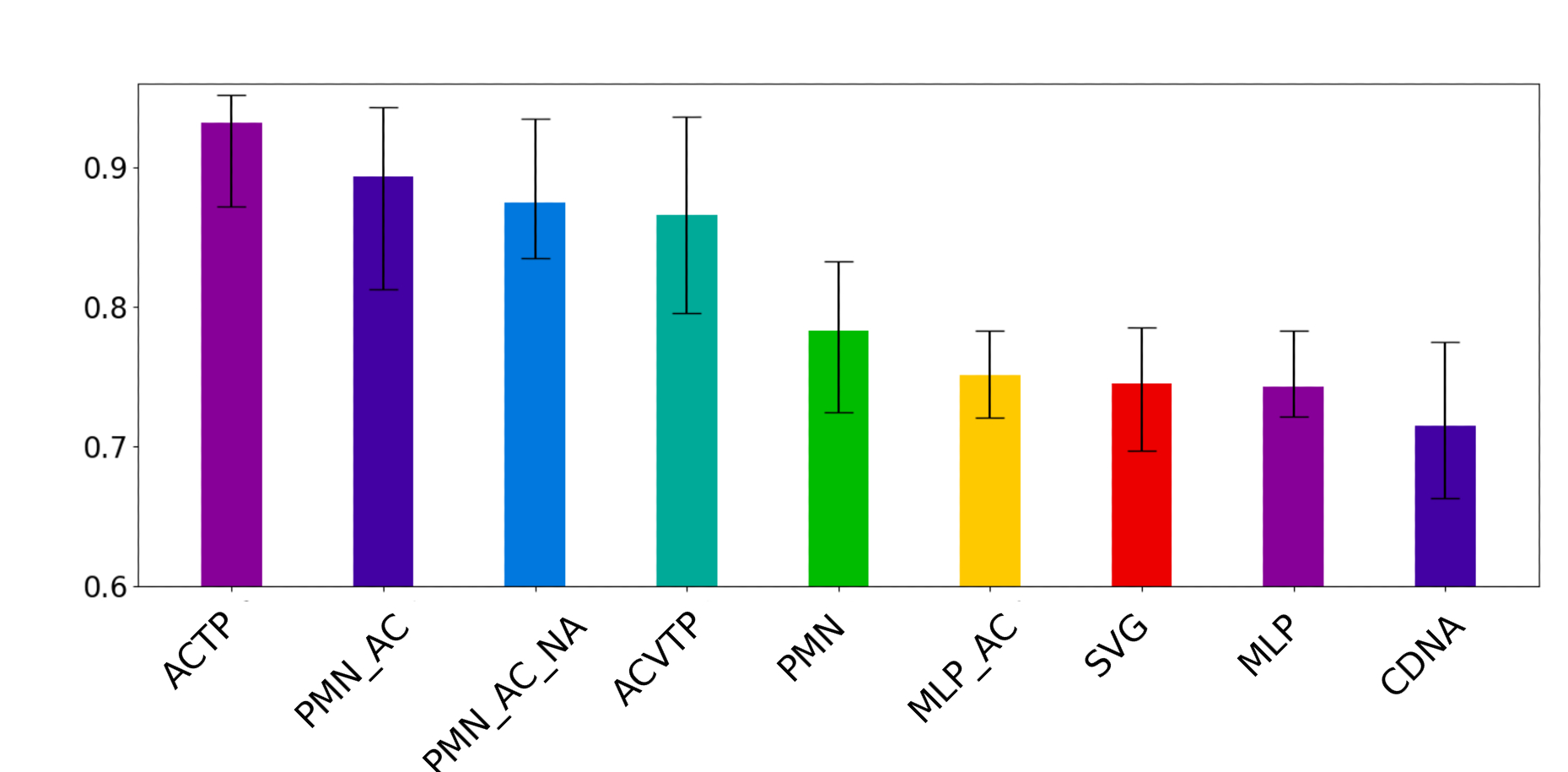}}%
        \caption{Slip Prediction score. Upper and lower variance values correspond to the test objects with highest and lowest scores.}
        \label{fig:pred}
    \end{figure}

    \begin{figure}[tb!]
        \centering
        \adjustbox{fbox= 0pt 1pt, frame}{\includegraphics[trim=0 0 0 10,clip, width=0.47\textwidth]{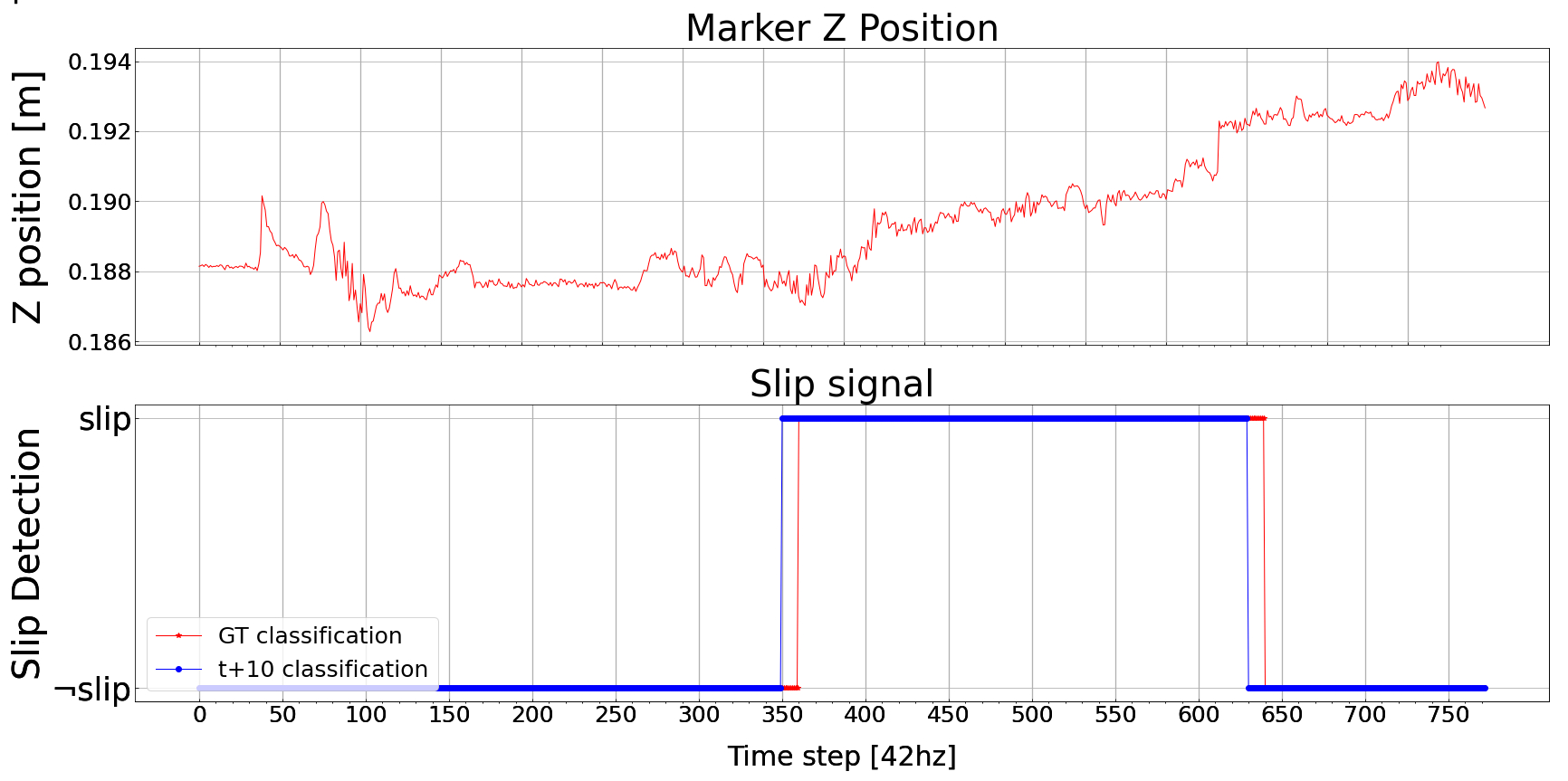}}
        \caption{Slip classification with ACTP model on GT and t+10 prediction signals.}
        \label{fig::classifier}
    \end{figure}

Fig.~\ref{fig::classifier} shows classification result for the ACTP model. The result shows the classification signal switches from non-slip to slip mode prior to GT classification as t+10 prediction signals capture the dynamic tactile changes 10 time steps ahead of the original signal. The classification signal in Fig.~\ref{fig::classifier} and the classification scores in Fig.~\ref{fig:pred} suggests the ACTP model yields the best slip prediction performance despite its poor MAE scores. While convolutional layers help ACTVP to achieve improved tactile prediction by spatio-temporal analysis, they increase the latent space dimensionality and it is possible that robot data fusion with tactile data happens less efficiently in ACTVP relative to its linear counterpart ACTP which has a smaller latent vector; Thus, better slip prediction in ACTP. The large variance of the slip score indicates SVG slip detection performance varies across different experimentation. According to slip score, PMN has the poorest performance.

The inference time for the best performing model (ACTP), using an AMD `Rhyzen threadripper 2950x’ 16 core processor, for tactile prediction and slip classification during robot motion (10 context and 10 prediction frames) was 14ms. This enables tactile predictions and predicted slip classification at 71Hz, higher than our control loop of 40Hz, we believe this is fast enough to react to predicted slip.


\section{Discussion and Conclusion}
\label{sec:conclusion}
    We presented two novel data-driven predictive models for tactile signals during real-world physical robot interaction tasks. We use a magnetic-based tactile sensor, the Xela uSkin, known for being difficult to analyse due to it's calibration challenges. We created a dataset of kinesthetically driven, teleoperated, pick and move tasks of flat-surfaced household objects and recorded the tactile sensation, proprioception robot data and the pose of the object relative to the robot's wrist. The data from different sources is synchronised. 

    We propose two novel data-driven predictive models trained on the dataset: (1) Action conditioned tactile prediction (ACTP) and (2) Action conditioned video tactile prediction (ACTVP). ACTP and ACTVP use different representations of the tactile data. We compare these models to state of the art video prediction model SVG~\cite{denton2018stochastic}, video prediction model CDNA~\cite{finn2016unsupervised} which has previously been applied to tactile prediction and the only existing bespoke tactile prediction model, PixelMotionNet~\cite{zhang2020towards}. We adjust PixelMotionNet to include action conditioning and remove the optical flow layer to enable insight into the effect of these two features. We show that our presented model ACTVP had the best performance metrics. However, qualitative analysis and the slip prediction task show that ACTP is the best performing model. We find that optical flow and encoder/decoder methods result in reduced prediction performance. CDNA's specific method of optical flow, where a network generates masks and applies kernels to these masks, results in poor performance despite previous success in video and video-based tactile prediction with physical robot interaction tasks. Qualitative analysis and the application task show quantitative metrics like MAE do not provide insight into practical model performance for specific application domains. For instance, a model that performs best for slip prediction may perform poorly for pushing tasks or vice versa.

    Due to the low resolution of the magnetic tactile sensor, it is unlikely the discussed pipeline will be capable of accurate prediction with more complex surfaces. However, higher resolution vision based tactile sensors may provide prediction models with a more accurate sense of touch, reducing uncertainty in the prediction models. With vision based sensors and more complex surfaces, ACTVP's convolutional architecture may produce more accurate results in comparison to ACTP. Likewise, the use of optical flow techniques may have increased benefit, especially CDNA, as the assumption made with respect to object masks may be more applicable in this setting. The inclusion of complex surface topologies with magnetic-based tactile sensors will increase the number of latent variables due to the tactile resolution. Learned priors, which attempt to estimate realistic values for these may have a positive impact.

    We use these models and combine them with a slip classification method to perform action-conditioned slip prediction. In contrast to the existing slip prediction methods (that train a single model for predicting slip only for a fixed horizon), our novel approach yields a more robust and reliable slip prediction framework in real-world manipulation tasks. In particular, our approach allows us to readily change the slip prediction horizon on the fly without retraining. We see space for the continued development of tactile prediction networks by introducing vision based tactile sensors for non-flat surfaces and integrating a multi-modal approach using visual features of the object to provide better context to the prediction models. Our future works also include integrating the developed tactile prediction/slip prediction methods in two different application domains (1) controlling a manipulator to avoid predicted slip and (2) robotic pushing.

\section*{Acknowledgements}
This work was partially supported by Centre for Doctoral Training, United Kingdom (CDT) in Agri-Food Robotics (AgriFoRwArdS) Grant reference: EP/S023917/1; Lincoln Agri-Robotics (LAR) funded by Research England; and by ARTEMIS project funded by Cancer Research UK C24524/A300038.

\bibliographystyle{plainnat}
\bibliography{main}

\end{document}